\documentclass{article}
\usepackage{ijcai23}

\usepackage{times}
\usepackage{soul}
\usepackage{url}
\usepackage[hidelinks]{hyperref}
\usepackage[utf8]{inputenc}
\usepackage[small]{caption}
\usepackage{graphicx}
\usepackage{amsmath}
\usepackage{amsthm}
\usepackage{booktabs}
\usepackage{algorithm}
\usepackage{algorithmic}
\usepackage[switch]{lineno}
\usepackage{color}
\usepackage{amssymb}
\usepackage{pifont}
\usepackage{subfigure}
\usepackage{xspace}
\usepackage{url}
\usepackage{bm}
\newcommand{\citet}[1]{\citeauthor{#1}~\shortcite{#1}}
\newcommand{\ie}{\emph{i.e.,}\xspace}
\newcommand{\aka}{\emph{a.k.a.,}\xspace}
\newcommand{\eg}{\emph{e.g.,}\xspace}

\title{Diffusion Models for Non-autoregressive Text Generation: A Survey}

\author{
Yifan Li$^1$
\and
Kun Zhou$^{2,3}$\and
Wayne Xin Zhao $^{1,3}$ \thanks{Corresponding Author.} \And
Ji-Rong Wen$^{1, 2, 3}$
\affiliations
$^1$Gaoling School of Artificial Intelligence, Renmin University of China\\
$^2$School of Information, Renmin University of China\\
$^3$Beijing Key Laboratory of Big Data Management and Analysis Methods
\emails
\{liyifan0925, batmanfly\}@gmail.com,
francis\_kun\_zhou@163.com,
jrwen@ruc.edu.cn
}

\begin{document}

\maketitle

\begin{abstract}
Non-autoregressive~(NAR) text generation has attracted much attention in the field of natural language processing, which greatly reduces the inference latency but has to sacrifice the generation accuracy. 
Recently, diffusion models, a class of latent variable generative models, have been introduced into NAR text generation, showing an improved text generation quality.
In this survey, we review the recent progress in diffusion models for NAR text generation. 
As the background,  we first present the general definition of diffusion models and the text diffusion models, and then discuss their merits for NAR generation.
As the core content, we further introduce two mainstream diffusion models in existing work of text diffusion, and review the key designs of the diffusion process. 
Moreover, we discuss the utilization of pre-trained language models (PLMs) for text diffusion models and introduce   optimization techniques for text data. 
Finally, we discuss several promising  directions and conclude this paper. Our survey aims to provide researchers with a systematic reference of related research on text diffusion models for NAR generation. We present our collection of text diffusion models at \url{https://github.com/RUCAIBox/Awesome-Text-Diffusion-Models}.

\end{abstract}

\begin{table*}[htb]
	\centering
	\begin{tabular}{l|clllccc}
		\toprule
  Model & NAR & Diffusion space & Noise schedule & Tasks & $x_0$-param & PLMs & Clamping\\
  \hline
  D3PM~\shortcite{DBLP:conf/nips/AustinJHTB21}                     & \ding{51} & Discrete & Mutual information   & UCG       & \ding{51} & \ding{55} & \ding{55} \\
  Diffusion-LM~\shortcite{lidiffusion}     & \ding{51} & Continuous & Sqrt                 & A2T       & \ding{51} & \ding{55} & \ding{51} \\
  Diffuseq~\shortcite{DBLP:journals/corr/abs-2210-08933}         & \ding{51} & Continuous & Sqrt                 & T2T       & \ding{51} & \ding{55} & \ding{51} \\
  SED~\shortcite{DBLP:journals/corr/abs-2211-04236}              & \ding{51} & Continuous & Cosine               & UCG, A2T  & \ding{51} & \ding{55} & \ding{55} \\ 
  SSD-LM~\shortcite{DBLP:journals/corr/abs-2210-17432}           & \ding{51} & Continuous & Cosine               & UCG, A2T  & \ding{51} & \ding{51} & \ding{55} \\
  DiffusionBERT~\shortcite{DBLP:journals/corr/abs-2211-15029}    & \ding{51} & Discrete & Spindle              & UCG       & \ding{51} & \ding{51} & \ding{55} \\
  CDCD~\shortcite{DBLP:journals/corr/abs-2211-15089}             & \ding{51} & Continuous & -                    & T2T       & \ding{51} & \ding{55} & \ding{55} \\
  Difformer~\shortcite{DBLP:journals/corr/abs-2212-09412}        & \ding{51} & Continuous & Linear               & T2T       & \ding{51} & \ding{51} & \ding{55} \\
  LD4LG~\shortcite{DBLP:journals/corr/abs-2212-09462}            & \ding{55} & Continuous & Linear               & UCG, A2T  & \ding{51} & \ding{51} & \ding{55} \\
  SeqDiffuSeq~\shortcite{DBLP:journals/corr/abs-2212-10325}      & \ding{51} & Continuous & Adaptive             & T2T       & \ding{51} & \ding{55} & \ding{55} \\
  Diff-Glat~\shortcite{Diffglat}      & \ding{51} & Discrete & -            & T2T       & \ding{51} & \ding{55} & \ding{55} \\
  GENIE~\shortcite{DBLP:journals/corr/abs-2212-11685}            & \ding{51} & Continuous & -                    & T2T       & \ding{55} & \ding{55} & \ding{51} \\
  DINOISER~\shortcite{DINOSIER}            & \ding{51} & Continuous & Linear                    & T2T       & \ding{51} & \ding{55} & \ding{55} \\
GlyphDiffusion~\shortcite{li2023glyphdiffusion}            & \ding{51} & Continuous  & -                    & T2T       & \ding{55} & \ding{55} & \ding{55} \\
  Diffusion-NAT~\shortcite{zhou2023diffusionnat}            & \ding{51} & Discrete & Linear                    & T2T       & \ding{51} & \ding{51} & \ding{55} \\
		\bottomrule   
	\end{tabular}
	\caption{Comparison of existing text diffusion models. $x_0$-param, PLMs and Clamping denote using the $x_0$-parameterized loss, PLMs, and the clamping trick, respectively. \textbf{UCG}, \textbf{A2T}, \textbf{T2T} refers to \textbf{U}n\textbf{C}onditional \textbf{G}eneration, \textbf{A}ttribute-\textbf{T}o-\textbf{T}ext and \textbf{T}ext-\textbf{T}o-\textbf{T}ext, respectively.}
	\label{tab:intro}
\end{table*}

~\section{Introduction}
Text generation~\cite{DBLP:journals/jair/GattK18} (\aka natural language generation) 
aims to generate human-like text  (\ie a sequence of word tokens) given the input data (\eg  sentence or keywords), enabling a wide range of real-world applications such as machine translation~\cite{DBLP:journals/corr/BahdanauCB14} and text summarization~\cite{DBLP:conf/aaai/NallapatiZZ17}.
Due to the excellent sequence  modeling capacity, deep learning has become the mainstream approach to developing the backbone for text generation models, exemplified by RNN~\cite{DBLP:conf/emnlp/ChoMGBBSB14} and transformer~\cite{DBLP:conf/nips/VaswaniSPUJGKP17}.
More recently, pre-trained language models~(PLMs)~\cite{li2021pretrained,LLMSurvey}  further raise the performance bar of text generation.
After being pre-trained on the large-scale general corpus, PLMs can be effectively fine-tuned for downstream tasks, leveraging the pre-learned rich knowledge to improve task performance.
Generally, existing  text generation methods mostly adopt the autoregressive way (AR) that generates the output tokens one by one.
Such a way is able to capture the sequential dependency relations among tokens, but would be time-consuming when generating long texts.
Thus, non-autoregressive (NAR) generation methods, which generate all tokens in parallel and greatly reduce the inference latency, have been proposed~\cite{DBLP:conf/iclr/Gu0XLS18}.

However, NAR models generally underperform AR ones on text generation accuracy, since the token dependency relations cannot be well captured by the parallel generation.
To narrow the performance gap, previous works have proposed various improvement techniques for NAR methods, \eg knowledge distillation~\cite{DBLP:conf/iclr/ZhouGN20} and large-scale pre-training~\cite{DBLP:conf/icml/QiG0YCLTLCZ0D21}.
More recently, diffusion models~\cite{DBLP:conf/icml/Sohl-DicksteinW15,DBLP:conf/nips/HoJA20}, a class of generative models that have shown superiority in image generation, are introduced into NAR text generation. 
In essence, diffusion models perform a multi-step denoising process to progressively convert a random noise into a data sample.
To adapt to NAR text generation tasks, diffusion models iteratively refine the intermediate generated results conditioned on the input data, which are shown to be potentially more capable of handling complex control conditions in producing high-quality  text~\cite{lidiffusion}. 
Further, by designing  proper sampling acceleration methods~\cite{DBLP:conf/iclr/SongME21}, diffusion models can well balance the inference latency  and generation quality, leading to an improved generation ability.

Existing studies have explored  two types of representative diffusion processes from image generation  into text generation, \ie \emph{continuous diffusion}~\cite{lidiffusion} and \emph{discrete diffusion}~\cite{DBLP:conf/nips/HoogeboomNJFW21,DBLP:conf/nips/AustinJHTB21} that perform the diffusion process in continuous latent representations  and discrete text tokens, respectively. We provide a detailed illustration of these studies and the major features  in Table \ref{tab:intro}.
However, due to the discrete essence and complex semantics of texts, it is not easy to effectively adapt the above diffusion models to NAR text generation tasks. 
Prior studies have introduced or devised specific strategies to improve the original settings of diffusion models for NAR text generation, including revising the training objective, adopting noise schedules tailored for text, and integrating PLMs.  
Despite this  progress, the field of diffusion models for text generation is still nascent, requiring a deep investigation in this line. 
For this purpose, a comprehensive survey that summarizes the recent  advances  is highly needed. 
To the best of our knowledge, this survey is the first literature review  that concentrates on the research of diffusion models for NAR text generation.

To start with, we first briefly review diffusion models, formulate text diffusion models and describe the connections between text diffusion models and NAR models in Section \ref{sec: overview}. 
Then, 
we introduce two mainstream diffusion models used for NAR text generation in Section \ref{sec: customized process}, and review four key designs of the diffusion process in Section \ref{sec: denoising network}, \ie denoising network, noise schedule, objective function and conditioning strategy. 
Next, we discuss the utilization of pre-trained language models (PLMs) for text diffusion in Section \ref{sec: PLMs} and present other optimizations of diffusion models for text data in Section \ref{sec: optimization}. 
Finally, we prospect the future directions and conclude this survey in Section \ref{sec: future directions}.

~\section{Overview of Text Diffusion Models} \label{sec: overview}

Diffusion models have made remarkable progress in generating continuous data, \eg image~\cite{DBLP:conf/nips/DhariwalN21} and audio~\cite{DBLP:conf/iclr/KongPHZC21}. Recently, their applications in discrete text data, referred to as \emph{text diffusion models}, are gaining growing attention. 
In this section, we first present a brief overview of the typical diffusion model~\cite{DBLP:conf/nips/HoJA20}, then give a formulated definition of text diffusion models,  and finally compare them with traditional NAR models for text generation.

~\subsection{Diffusion Models} \label{subsec: diffusion models}
Diffusion models are a class of latent variable models characterized by a forward and a reverse Markov process. 
The forward process $q(x_t|x_{t-1})$ gradually corrupts the data sample $x_0$ using random noise. 
The reverse process $p_{\theta}(x_{t-1}|x_t)$ relies on a denoising network $f_{\theta}$ to progressively recover a random noise into the desired data sample.

To be more specific, given a data sample $x_0\sim q(x)$, the forward process generates a sequence of latent variables $x_1, ..., x_T$ by sampling from 
\begin{equation}\label{equ:3}
    q(x_t|x_{t-1})= \mathcal{N}(x_t; \sqrt{1-\beta_t}x_{t-1}, \beta_t\mathbf I),
\end{equation}
 where $\beta_t\in(0,1)$ is the noise scale. 
 Following a pre-defined noise schedule, $\beta_t$ increases as the timestep grows and eventually corrupts $x_0$ into a random noise. Then, based on the reparameterization trick, arbitrary intermediate latent variable $x_t$ can be sampled from $x_0$ in a closed form:
\begin{equation}
    q(x_t|x_0) = \mathcal{N}(x_t; \sqrt{\bar\alpha_t}x_0, \sqrt{1-\bar\alpha_t}\mathbf I),
\end{equation}
where $\alpha_t=1-\beta_t$ and $\bar{\alpha}_t=\prod_{i=1}^t \alpha_i$. 
The reverse process is the approximation of the posterior $q(x_{t-1}|x_t)$, which can be seen as a Gaussian when $\beta_t$ is small enough. In this way, the reverse process is also formulated as a Gaussian distribution:
\begin{equation} \label{equ: reverse}
    p_{\theta}(x_{t-1}|x_t)=\mathcal{N}(x_{t-1}; \mu_{\theta}(x_t, t), \Sigma_{\theta}(x_t, t)), 
\end{equation}
where $\mu_{\theta}(x_t, t)$ and $\Sigma_{\theta}(x_t, t)$ are parameterized by a denoising networks $f_{\theta}$ like U-Net~\cite{DBLP:conf/miccai/RonnebergerFB15} or transformer~\cite{DBLP:conf/nips/VaswaniSPUJGKP17}. During inference, the reverse process begins with sampling noise from a Gaussian distribution $p(x_T)=\mathcal{N}(x_T;0, \mathbf{I})$ and iteratively denoise it by $p_{\theta}(x_{t-1}|x_t)$ until obtaining $x_0$.
The learning objective of diffusion models is derived from the variational lower bound of the negative log-likelihood of input $x_0$, denoted as:
\begin{align} \label{equ: vlb}
    \begin{aligned}
    \mathcal L_{\rm vlb}= & \mathbb{E}_q[\underbrace{D_{\rm KL}(q(x_t|x_0)||p_\theta(x_T))}_{\mathcal L_T}]-\underbrace{\log p_\theta(x_0|x_1)}_{\mathcal L_0}\\    & +\mathbb{E}_q[\sum_{t=2}^T\underbrace{D_{\rm KL}(q(x_{t-1}|x_{t}, x_0)||p_\theta( x_{t-1}|x_t))}_{\mathcal L_{t-1}}].
    \end{aligned}
\end{align}
The final training objective is derived from $\mathcal L_{t-1}$. With additional condition on $x_0$, the posterior of the forward process $q(x_{t-1}|x_t, x_0)$ can be calculated using Bayes theorem, then the simplified objective $\mathcal{L}_{\rm simple}$ can be expressed as:
\begin{equation} \label{equ: simple loss}
    \mathcal L_{\rm simple}=\sum_{t=1}^T\mathbb{E}_q \big[||\mu_t(x_t, x_0)-\mu_{\theta}(x_t, t)||^2 \big], 
\end{equation}
where $\mu_t$ is the mean of posterior $q(x_{t-1}|x_t, x_0)$. Therefore, the denoising network $f_{\theta}$ is trained to predict $\mu_t$ given $x_t$ and $t$. Through different parameterization strategies, the prediction objective can also be  the noise 
$\epsilon_t$~\cite{DBLP:conf/nips/HoJA20} or original data $x_0$~\cite{lidiffusion}.

~\subsection{Text Diffusion Models} \label{subsec: text diffusion models}

Text diffusion models aim to gradually recover a random noise to a desired text based on the given input data. 
The starting noise can be discrete (\eg [\textsc{Mask}] tokens) or continuous (\eg random Gaussian noise), corresponding to the discrete or continuous diffusion model (Section \ref{sec: customized process}).
The denoising process relies on a parameterized denoising network, which is generally implemented by the transformer architecture~\cite{DBLP:conf/nips/VaswaniSPUJGKP17}. 
During training, the denoising network learns to recover the intermediate noised results based on the settings of noise schedule, objective function and conditioning strategy (Section \ref{sec: denoising network}).
During inference, starting from a random noise $\mathcal{Y}_T$, the denoising network progressively denoises it at each step, until producing the target text. 
Note that at each step, following the NAR generation manner, text diffusion models predict all the latent variables in parallel.
The above process can be formulated as: 
\begin{equation}p(\mathcal{Y}|c)=\prod_{t=0}^{T-1}\prod_{i=1}^n p(y_i|\hat{\mathcal{Y}}_{t+1}, c, t),
\end{equation}
where $\mathcal{Y}$ is the target text consisting of a sequence of tokens $y_i$, $\hat{\mathcal{Y}}_{t+1}$ denotes the latent variables predicted at the $t+1$ timestep, $c$ is the input condition and $t$ denotes the timestep.

To improve the performance, it is significant to incorporate advanced NLP techniques with text diffusion models.
As an important progress in the NLP area, pre-trained language models~(PLMs) have been explored for integration with text diffusion models (Section \ref{sec: PLMs}). 
Moreover, a variety of optimization strategies have been proposed in existing text diffusion models to better capture the unique characteristics of text data (Section \ref{sec: optimization}).

~\subsection{Merits of Text Diffusion Models for NAR Generation}
As mentioned before, the parallel generation manner of NAR methods would greatly reduce the inference latency, but is incapable of learning the dependency relations among tokens, leading to a decreased accuracy.
While, text diffusion models have several merits that can help improve the NAR generation accuracy. 
In this part, we show three major merits of text diffusion models,  
\ie constrained iterative refinement, introducing intermediate control and trading off time-cost and quality.

\paragraph{Constrained Iterative Refinement.}
Typical NAR models generate all the target tokens in parallel, hence the inference latency would be rather smaller than in AR methods.
Therefore, existing works~\cite{DBLP:conf/emnlp/GhazvininejadLL19,DBLP:conf/nips/GuWZ19} also incorporate the iterative refinement strategy to enhance the quality of the generated results.
However, with the increase of the iteration steps, 
it also raises the problem of how to effectively control or supervise the intermediate refinement process on the discrete target tokens, restraining the improvement of the generation performance.
As a promising solution, text diffusion models provide a constrained iterative refinement process for gradually enhancing the generation quality, where each step is constrained to denoise a random noise with a pre-defined variation.

\paragraph{Imposing Intermediate Control.}
In the iteration process, it is also hard to directly control the intermediate results for existing NAR methods, especially for injecting complex controlled conditions (\eg following a syntax parse tree).
For text diffusion models, existing works have extensively studied injecting the control  conditions in the intermediate results, by adding extra classifiers~\cite{lidiffusion} or using classifier-free controls~\cite{DBLP:journals/corr/abs-2211-04236}.
As theoretically and empirically proved, these approaches can better steer the intermediate prediction steps toward the generation of the target text that satisfies the control requirements. 

\paragraph{Trading off between Time Cost and Quality.}
During inference, existing NAR methods seek to strike a balance between time cost and quality.
They mainly rely on tuning the iterative turns to achieve the goal, where decreasing the number of iterations would increase the inference speed but potentially sacrifice the generation quality.
To provide a more flexible trade-off between quality and inference time, text diffusion models can adopt inference acceleration techniques, \eg DDIM~\cite{DBLP:conf/iclr/SongME21}.
Empirical results have shown that these methods can flexibly adjust the iteration steps with a slight decrease in the generation quality, thereby achieving a better trade-off between cost and quality.

\begin{figure*}[htb]
	\centering
	\subfigure[Discrete text diffusion model.]{\label{fig: ddp}
		\centering
	\includegraphics[height=3.55cm]{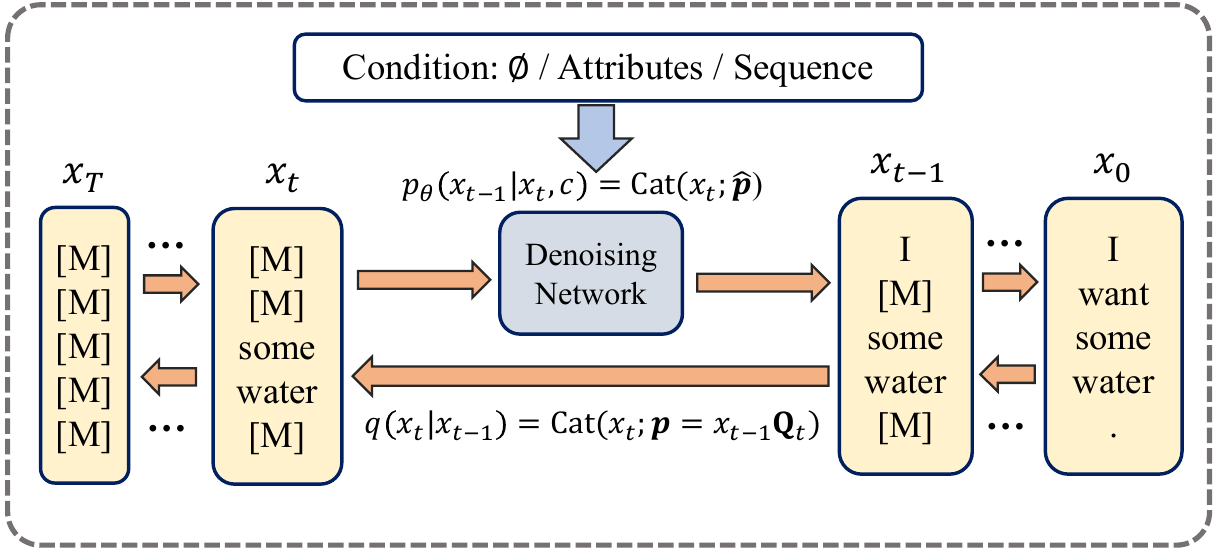}
	}\hspace{-2.5mm}
	\subfigure[Continuous text diffusion model.]{\label{fig: cdp}
		\centering
		\includegraphics[height=3.55cm]{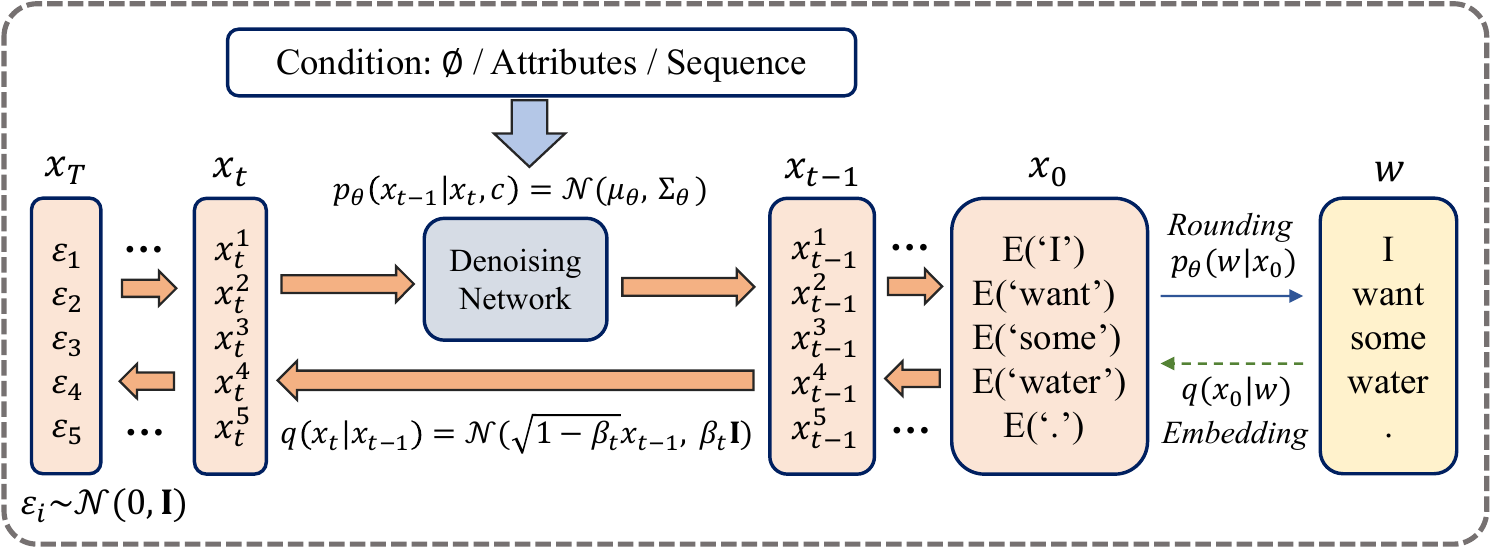}
	}
	\centering
	\caption{Illustrations about the discrete and continuous text diffusion model. In the discrete text diffusion model, ``[M]'' represents the [\textsc{Mask}] token. In the continuous text diffusion model, E($\cdot$) represents the embedding function.}
	\label{fig-training}
\end{figure*}

~\section{Customized Diffusion Models for Text} \label{sec: customized process}
The adaptation of diffusion models to NAR text generation is challenging due to the discrete nature of text. Specially, discrete tokens cannot be directly corrupted by continuous noise, so we need to design specific adaptions of typical diffusion models for text data. In this section, we review the recent progress of diffusion models tailored for text data, which perform either the diffusion process on discrete tokens or the continuous diffusion on latent representations of tokens (\eg word embeddings).
 
~\subsection{Discrete Text Diffusion Model}
The overview of the discrete text diffusion model is shown in Figure \ref{fig: ddp}. The diffusion process in the discrete domain is first introduced by \citet{DBLP:conf/icml/Sohl-DicksteinW15}, which proposes a binomial diffusion process to predict the binary representations of the continuous data. \citet{DBLP:conf/nips/HoogeboomNJFW21} further explore the diffusion process for discrete states with a uniform transition kernel. D3PM~\cite{DBLP:conf/nips/AustinJHTB21} proposes a general framework for diffusion models for discrete states and first tests discrete diffusion models on large-scale corpora. For discrete data  $x \in 1,..., K$, D3PM designs a transition matrix $[\mathbf Q_t]_{ij}=q(x_t=j|x_{t-1}=i)$ to corrupt $x$. The forward process in Eq.~\eqref{equ:3} now takes the following  form:
\begin{equation}
    q(x_t| x_{t-1})={\rm Cat}( x_t; \bm p= x_{t-1}\mathbf Q_t),
\end{equation}
where $x$ is represented by one-hot vectors and ${\rm Cat}(\cdot)$ is a categorical distribution over $x$. Following a similar derivation in the continuous diffusion process, $ x_t$ can be sampled by:
\begin{equation}
    q( x_t| x_0)={\rm Cat}( x_t; \bm p= x_0\bar{\mathbf Q}_t ),
\end{equation}
where $\bar{\mathbf Q}_t=\prod_{i=1}^t \mathbf{Q}_i$. By using Bayes theorem, the posterior $q(x_{t-1}| x_{t},  x_0)$ can be written as:
\begin{equation}
    q( x_{t-1}| x_{t},  x_0)={\rm Cat}\big( x_{t-1};\bm p=\frac{ x_t{\mathbf Q_t^{\top}}\odot  x_0\bar{\mathbf{ Q}}_{t-1}}{ x_0\bar{\mathbf{ Q}}_{t} x_t^{\top}}\big),
\end{equation}
where ``$\odot$'' is an element-wise multiplication. Then $\mathcal{L}_{\rm vlb}$ can be calculated by accumulating the KL-divergence between every component of $q$ and $p_{\theta}$, following Eq.~\eqref{equ: vlb}.

By designing different transition matrices $\mathbf{Q}_t$, the above framework can incorporate specific diffusion processes to generate text data. D3PM introduces a transition matrix with an absorbing state, allowing each token to be transformed into a [\textsc{Mask}] token with a probability $\beta_t$. During inference, D3PM starts from a sequence  of full [\textsc{Mask}] tokens and iteratively substitutes [\textsc{Mask}] with word tokens until  the desired text is generated.

~\subsection{Continuous Text Diffusion Model}
The overview of the continuous text diffusion model is illustrated in Figure \ref{fig: cdp}, where discrete tokens are first mapped to embeddings before  the continuous diffusion process. Diffusion-LM~\cite{lidiffusion} first applies continuous diffusion models to text generation, and adds  an embedding step $q_{\phi}( x_0| w)=\mathcal N({\rm EMB}( w), \sigma_0\mathbf{I})$ to the forward process, where ${\rm EMB}( w)$  is a randomly initialized embedding function that projects discrete tokens $ w$ to continuous space. For the reverse process, Diffusion-LM incorporates a rounding step $p_{\theta}( w| x_0)=\prod_{i=1}^n p_{\theta}(w_i|x_0)$ to map the final generation results to discrete tokens, where $p_{\theta}(w_i|x_i)$ is a softmax function. The inference process starts from a random noise and follows the typical continuous diffusion process in Section~\ref{subsec: diffusion models} to recover the noise to word embeddings, which are finally mapped to word tokens through the rounding step. To jointly learn the denoising network and the mapping relationship between embeddings and word tokens, Diffusion-LM reformulates the training objective in Eq.~\eqref{equ: vlb}:
\begin{equation}
        \mathcal L_{\rm vlb}^{'} = \mathbb E_q[\mathcal L_{\rm vlb} + \log q_{\phi}( x_0| w) - \log p_{\theta}( w| x_0)],
\end{equation}
which can be further simplified as:
\begin{align}
    \begin{aligned}
    \mathcal L_{\rm simple}^{'} = & \mathbb E_q[\mathcal L_{\rm simple} + ||{\rm EMB}( w)-\mu_{\theta}( x_1, t_1)||^2 \\ &- \log p_{\theta}( w| x_0)].
     \end{aligned}
\end{align}

Different from the above works which map tokens to word embeddings, SSD-LM~\cite{DBLP:journals/corr/abs-2210-17432} utilizes a simplex representation over the vocabulary $V$ to represent tokens. Given a token $w$, its simplex representation $\Tilde{\boldsymbol{w}}\in \{-K, +K\}^{|V|}$ can be expressed as:
\begin{equation} \label{equ: ssd}
   \Tilde{w}_{(i)}=
    \left\{
    \begin{array}{cc}+K
          \quad {\rm when}\quad w = V_{(i)} \\
         -K\quad {\rm when}\quad w \neq V_{(i)}
    \end{array}\right..
\end{equation}
During inference, SSD-LM starts with a random noise and iteratively denoises  it following the standard continuous diffusion, but adopts a logits-projection operation  converting  the logits to the similar almost-one-hot representations mentioned in Eq.~\eqref{equ: ssd} before  the next  decoding step.

Furthermore, GlyphDiffusion~\cite{li2023glyphdiffusion} renders the target text as a glyph graph and casts the text generation task into an image generation task, which can be naturally handled by the continuous diffusion model. Following popular settings in image diffusion models, GlyphDiffusion adopts a U-Net as the denoising network and predicts current noise with classifier-free guidance~\cite{ho2021classifier}. Its training objective can be written as: 

\begin{equation}
    \mathcal{L} = \mathbb E_{x_0, \epsilon, t}(||\epsilon-\hat{\epsilon}_{\theta}(x_t, c)||^2_2),
\end{equation}
where $\epsilon$ is the target noise and $c$ is the condition text.

~\section{Key Designs in Diffusion Process} \label{sec: denoising network}

The denoising network and the related   settings (\eg noise schedule, objective function and conditioning strategy) are crucial parts of text diffusion models and significantly impact the generation quality. In this section, we will introduce these designs and improvements for them in text diffusion models.

~\subsection{Denoising Network} \label{subsec: denoising network}
The denoising network aims to remove noise from intermediate results in the reverse process of diffusion models. Different from vision diffusion models which adopt U-Net~\cite{DBLP:conf/miccai/RonnebergerFB15} as denoising networks, text diffusion models typically  use transformer~\cite{DBLP:conf/nips/VaswaniSPUJGKP17} to better capture the dependency between tokens. 

\paragraph{Transformer.}
Transformers have dominated the field of natural language processing in recent years. A transformer is an encoder-decoder neural network composed of multiple transformer layers, with each including several feed-forward networks and multi-head self-attention functions $A(\cdot)$, which can be written as:
\begin{equation}
    A(x)={\rm softmax}(\frac{\mathbf{Q}\mathbf{K}^{\top}}{\sqrt{d}})\mathbf{V},
\end{equation}
where $x$ is the input sequence and is projected to $\mathbf{Q}$, $\mathbf{K}$ and $\mathbf{V}$ by different weight matrices and $d$ represents the hidden dimension. Due to their strong performance in text generation, most text diffusion models adopt the encoder or the whole part of transformers as the denoising network. Some studies  further utilize PLMs based on transformers (\eg BERT~\cite{DBLP:conf/naacl/DevlinCLT19} and RoBERTa~\cite{DBLP:journals/corr/abs-2210-17432}). 

~\subsection{Noise Schedule} \label{subsec: noise schedule}
The noise schedule $\beta$ is a function of the noise scale depending on the timestep, which controls the frequency of different input data of the denoising network. During the training stage, the original text is corrupted by a noise $\beta_t$ where $t$ is a randomly sampled timestep and fed into the denoising network. The noise schedule affects the denoising process for recovering the target text, thereby significantly impacting the generation quality. Existing methods either adopt popular noise schedules in vision tasks or devise new noise schedules tailored to the discrete nature of text data. 
\paragraph{Linear Schedule.}
DDPM~\cite{DBLP:conf/nips/HoJA20} proposes the linear schedule where $\beta_t$ ranges from $10^{-4}$ to $0.02$ linearly. 
Such a schedule ensures that the noise scale is relatively small at the beginning, making it easier for the denoising network to recover  the data, while eventually corrupting the original data to random noise by increasing the noise scale. Difformer~\cite{DBLP:journals/corr/abs-2212-09412} and LD4LG~\cite{DBLP:journals/corr/abs-2212-09462} follow a similar setting in text diffusion models. Based on the linear schedule, DINOISER~\cite{DINOSIER} further introduces noise scale clipping, which only samples noises whose scale is beyond a dynamic bound. Such a way enables the sampled noise to be sufficiently large to corrupt word embeddings and force the text diffusion model to better leverage source conditions.
\paragraph{Cosine Schedule.}
\citet{DBLP:conf/icml/NicholD21} argue that the noise scale in linear schedule increases more quickly than necessary, making the latent variables in the last quarter of the linear schedule almost purely noise. Thus, they propose the cosine schedule by defining $\bar\alpha_t=\frac{f(t)}{f(0)}$, where $f(t)=\cos(\frac{t/T+s}{1+s}\cdot \frac{\pi}{2})^2$. Cosine schedule slows down the growth rate of the noise scale and is adopted by \citet{DBLP:journals/corr/abs-2211-04236} and \citet{DBLP:journals/corr/abs-2210-17432} for NAR text generation.
\paragraph{Mutual Information Schedule.}
D3PM~\cite{DBLP:conf/nips/AustinJHTB21} designs the mutual information schedule for the discrete diffusion process by linearly interpolating the mutual information between the latent variables and the original data. Specifically, for discrete diffusion models with absorbing states, the schedule reduces to $\beta_t=(T-t+1)^{-1}$, which is the same as the schedule in \citet{DBLP:conf/icml/Sohl-DicksteinW15}.
\paragraph{Sqrt Schedule.}
Diffusion-LM~\cite{lidiffusion} observes that the nearest neighbors of words in the embedding space stay constant after corruption and attributes this phenomenon to the small initial noise scale in traditional schedules.
Thus, it introduces the sqrt schedule by defining $\bar\alpha_t=1-\sqrt{t/T+s}$,  which has a higher initial noise scale and increasing rate, while gradually slowing down to avoid producing too many highly corrupted latent variables. 
Similar methods have also been used in \citet{DBLP:journals/corr/abs-2210-08933}. 
\paragraph{Spindle Schedule.}
The easy-first policy~\cite{kasai2020non} for NAR text generation argues that common words should be generated first to serve as the context for the subsequent generation of rare  words. Therefore, DiffusionBERT~\cite{DBLP:journals/corr/abs-2211-15029} proposes the spindle schedule, which assigns higher corruption probabilities to  more informative tokens. 
As a result, rare tokens will be replaced by [\textsc{Mask}] tokens at the start of the forward process and recovered at the end of the denoising process.
\paragraph{Adaptive Schedule.}
\citet{DBLP:journals/corr/abs-2212-10325}  propose that the difficulty of predicting $x_0$ should increase linearly along with the timestep. To this end, they design an   adaptive schedule by learning the relationship between the noise scale and the loss from an existing schedule (\eg sqrt schedule). During training, the noise scale is updated based on the observed loss.

~\subsection{Objective Function} \label{subsec: objective function}
As another key part, 
we also need to adapt the objective function of the denoising network in diffusion models to text generation. 
For example, the original $\mu_t$-parameterized loss may not be the optimal choice for predicting word embeddings, and additional embedding and rounding steps in continuous text diffusion models also need extra loss terms. Further reparameterization can reduce the complexity of the original loss.
\paragraph{$x_0$-parameterized Loss.} As mentioned in Eq.~\eqref{equ: simple loss}, the training objective for typical diffusion models can be simplified as the prediction of $\mu_t$, the mean of the posterior $q(x_{t-1}|x_0, x_t)$. However, \citet{lidiffusion} find that this objective can cause the prediction of $x_0$ to fail to converge to any word embeddings,  because the denoising network lacks sufficient constraints for $x_0$ when predicting $\mu_t$. Therefore, they propose to parameterize the training objective by the original text $x_0$, which can be written as:
\begin{equation}
    \mathcal L_{\rm simple}=\sum_{t=1}^T\mathbb{E}_{q}[||f_{\theta}(x_t, t)-x_0||^2],
\end{equation}
where $f_{\theta}$ is the denoising network. By doing so, the training objectives of the diffusion model in every timestep can be unified into the same one that predicts $x_0$. Such a loss is widely adopted in follow-up text diffusion models.   

\paragraph{Auxiliary Loss.}
Since continuous text diffusion models need to ground  embeddings to word tokens during inference, 
\citet{lidiffusion} introduce a new term $\mathcal L_{\rm round}=-\log p_{\theta} ({w}|x_0)$ in the objective function to better learn the mapping relationship between $w$ and $x_0$, where $p_{\theta} ({w}|x_0)$ is a softmax distribution over the vocabulary. While this objective jointly trains the embedding and the  diffusion process, it tends to learn a shortcut solution where every embedding is close to each other, forming an anisotropic embedding space. 
In fact, as $x_0$ is produced by adding a small amount of noise into the token embedding of $w$, it would be easily predicted and fails to provide enough guidance for the model training. 
Therefore, \citet{DBLP:journals/corr/abs-2212-09412} propose $\mathcal L_{\rm anchor}=-\log p_{\theta} ({w}|\hat{x}_0)$ to substitute $\mathcal L_{\rm round}$, where $\hat{x}_0$ represents the model prediction of $x_0$.  Larger distances between  $\hat{x}_0$ and ${w}$ ensure that the loss can offer sufficient guidance to regularize learned embeddings.

\paragraph{Surrogate Loss.} RDM~\cite{DBLP:journals/corr/abs-2302-05737} proposes to reparameterize the denoising process of discrete diffusion models by introducing step-wise routing indicators $\bm v_t = [v_t^{(1)}, v_t^{(2)}]$, where $v_t^{(1)}\sim {\rm Bernouli}(\lambda_t^{(1)})$ selects noisy tokens for recovering and $v_t^{(2)}\sim{\rm Bernouli}(\lambda_t^{(2)})$ chooses denoised tokens for corruption. Under this parameterization method, the training objective at the $t$-th timestep can be reformulated to:
\begin{equation}
    \mathcal{L}_t = \mathbb{E}[-\lambda_{t-1}^{(2)}\sum_{n=1}^N(1-b_{t, n})x_{0, n}^{\top}\log f(x_{t, n}; \theta)],
\end{equation}
where $b_t = 1$ if $x_t=x_0$ and else $0$, and $f(x_t, \theta)$ is the predicted $x_0$ by the denoising network. In this way, the training objective of RDM can be implemented by a multi-class cross-entropy loss.

~\subsection{Conditioning Strategy} \label{subsec: conditioning strategy}
By setting different conditions $c$ mentioned in Section \ref{subsec: text diffusion models}, text generation tasks can be further categorized into unconditional generation, attribute-to-text generation (\eg topic control) and text-to-text generation (\eg machine translation). Existing text diffusion models design various conditioning strategies to incorporate different conditions $c$ with denoising networks for these tasks. In this section, we will discuss these conditioning strategies.

\paragraph{Unconditional Generation.}
When setting $c$ to  empty, the task becomes unconditional text generation, where random noises are transformed to text sequences through the reverse process without other constraints. SED~\cite{DBLP:journals/corr/abs-2211-04236} and DiffusionBERT~\cite{DBLP:journals/corr/abs-2211-15029} follow such a task setting to evaluate the basic text generation ability of text diffusion models. 
 
\paragraph{Attribute-to-text Generation.}
When setting $c$ to attributes such as topic or sentiment, the task becomes attribute-to-text generation.
A classic method for this task is classifier-guidance, which adopts an existing classifier to provide gradient information as guidance during the generation process.  Diffusion-LM~\cite{lidiffusion} focuses on fine-grained control  conditions, where $c$ could be semantic contents (\eg five-star rating) or a syntactic parse tree. Following the plug-and-play method~\cite{DBLP:conf/iclr/DathathriMLHFMY20}, Diffusion-LM does not incorporate  conditions directly into the denoising network. Instead, they introduce an extra attribute classifier to guide the generation results at inference time. Similar to the classifier-guided vision diffusion models~\cite{DBLP:conf/nips/DhariwalN21}, Diffusion-LM runs gradient updates on the intermediate result during inference. The reverse process in Eq.~\eqref{equ: reverse} can be reformulated as:
\begin{equation}
    p_{\theta}({x}_{t-1}|{x}_t)=\mathcal{N}(x_{t-1}; \mu_{\theta}+s\nabla \log p(c|x_{t-1}), \Sigma_{\theta}),
\end{equation}
where $\nabla \log p(c|x_{t-1})$ is the gradient from the classifier and $s$ is the gradient scale.
Another alternative  approach, classifier-free~\cite{ho2021classifier},  explicitly introduces condition information into the denoising network.
LD4LG~\cite{DBLP:journals/corr/abs-2212-09462} uses class embeddings as conditions, which are combined with the latent variables via a cross-attention layer in the denoising network. Similarly, SED~\cite{DBLP:journals/corr/abs-2211-04236} uses condition embeddings during training, but combines them with the latent variables through the self-conditioning technique, which will be further discussed in Section~\ref{sec: optimization}.

\paragraph{Text-to-text Generation.}
When setting $c$ to text such as  token sequences or passages, the task becomes text-to-text generation, \eg  machine translation and text summarization. These tasks are typically more difficult than attribute-to-text tasks, as they cannot be well controlled by simple attribute classifiers. Therefore, classifier-guidance methods are no longer a suitable choice. Diffuseq~\cite{DBLP:journals/corr/abs-2210-08933} proposes the partially noising strategy to incorporate the condition text with the continuous diffusion process. Specifically, it concatenates the condition $c$ and the target sequence as the input of the denoising network. 
During the forward process, the concatenated sequence is partially corrupted by adding noise only to the target sequence while keeping $c$ unchanged. The reverse process begins with the concatenation of $c$ and random noise, with the condition  $c$ unchanged during inference.
\citet{DBLP:journals/corr/abs-2212-09412} and \citet{DBLP:journals/corr/abs-2212-10325} propose to use an integral transformer as the denoising network. The encoder is used to generate embedding representations of $c$,  and the decoder combines the corrupted target sequence embedding and the embedding of $c$ via cross-attention to predict the original target sequence. 

~\section{Utilization of Pre-trained Language Models} \label{sec: PLMs}
Since PLMs have achieved remarkable performance on various text generation tasks, we can integrate PLMs into text diffusion models to improve generation performance.  
In this section, we will introduce existing works that incorporate PLMs with text diffusion models.
~\subsection{PLMs as Denoising Networks}
The denoising network of discrete diffusion models aims to recover the original sequence corrupted by [\textsc{Mask}] tokens, which is similar to the pre-training tasks of existing PLMs, \eg masked language model~\cite{DBLP:conf/naacl/DevlinCLT19}. Therefore, it is promising to adopt PLMs as the denoising network in discrete text diffusion models. 
DiffusionBERT~\cite{DBLP:journals/corr/abs-2211-15029} uses a pre-trained BERT as the denoising network. However, the denoising network is typically conditioned  on timesteps for prediction, while PLMs do not encounter the timestep information during pre-training. To tackle this issue, DiffusionBERT introduces time-agnostic decoding, which does not explicitly incorporate timestep information in prediction but lets the model infer it based on the number of corrupted tokens.
 To better adapt text diffusion models to conditional text generation tasks, Diffusion-NAT~\cite{zhou2023diffusionnat} adopts a pre-trained BART~\cite{DBLP:conf/acl/LewisLGGMLSZ20} with a revised NAR decoding process as the denoising network. The condition and the corrupted text are fed to the encoder and decoder of BART, respectively. As for continuous text diffusion models,  SSD-LM~\cite{DBLP:journals/corr/abs-2210-17432} utilizes a pre-trained RoBERTa~\cite{DBLP:journals/corr/abs-1907-11692} as the denoising network to accelerate the convergence.
~\subsection{Diffusion on PLM's  Latent Space}
Latent diffusion models~\cite{DBLP:conf/cvpr/RombachBLEO22} conduct the diffusion process in the latent space of a pre-trained image autoencoder and achieve impressive results in the task of text-guided image generation. Similar approaches are also applied in text diffusion models. LD4LG~\cite{DBLP:journals/corr/abs-2212-09462} learns a text diffusion model in the latent space of a pre-trained BART~\cite{DBLP:conf/acl/LewisLGGMLSZ20}. During training, the BART encoder converts the text to embeddings, which are then corrupted and recovered through a continuous diffusion process. During inference, the denoising network recovers embeddings from random noise, which are then decoded into text by the BART decoder. LatentOps~\cite{DBLP:journals/corr/abs-2208-00638} utilizes a pre-trained GPT-2~\cite{radford2019language} to map latent vectors obtained by an ODE sampler back to the discrete text space. Difformer~\cite{DBLP:journals/corr/abs-2212-09412} initializes the embeddings from \texttt{bert-base} and \texttt{bert-base-multilingual} for text summarization and machine translation respectively.

~\subsection{Revising Pre-training Tasks of PLMs}
Although PLMs can offer an effective model initialization and accelerate convergence for text diffusion models, the latent space of PLMs may be a sub-optimal choice for diffusion models due to the discrepancy in training objectives. 
Thus, several recent works focus on revising the pre-training tasks of PLMs and pre-train new PLMs specifically for text diffusion models.
GENIE~\cite{DBLP:journals/corr/abs-2212-11685} designs a pre-training task similar to the masked language model task, called continuous paragraph denoising (CPD) and pre-trains a text diffusion model from scratch. Given a document $d$, CPD replaces a paragraph $p$ in it with [\textsc{Mask}] and passes the corrupted document $d'$ through an encoder to obtain a context representation. Then the paragraph $p$ is corrupted and fed into the denoising network along with the context representation to predict the noise added during the diffusion process.

~\section{Other Improvement Strategies for Text Data} \label{sec: optimization}
In addition to the aforementioned methods, there are many other techniques to further improve the performance of text diffusion models, which are designed for unique characteristics of text data or borrowed from diffusion models in other fields. In this section, we will  review these methods.

\paragraph{Clamping Trick.}
Diffusion-LM~\cite{lidiffusion} clamps the prediction result of the denoising network $f_{\theta}(x_t,t)$ to the nearest word embedding during inference before using it for the next prediction. This forces $f_{\theta}(x_t,t)$ to be concentrated on real words in the vocabulary, resulting in a lower rounding loss. However, the clamping trick needs to calculate the distance between every word embedding and the prediction result, which can lead to high computational costs when applied at every inference step. Considering this issue, GENIE~\cite{DBLP:journals/corr/abs-2212-11685} only applies the clamping trick at the end of inference.

\paragraph{Self-conditioning.}
In the reverse process of standard  diffusion models, the denoising network makes predictions based on the current latent variables $x_t$ and timestep $t$. 
Analog Bits~\cite{DBLP:journals/corr/abs-2208-04202} further includes the previous prediction of the original data $\tilde{x}_0$ as the input to the denoising network, which is referred to as self-conditioning. In practice, self-conditioning takes the concatenation of $x_t$ and $\tilde{x}_0$ as the input to the denoising network. 
However, during training, $\tilde{x}_0$ is not available as it is in the inference stage.
Therefore, Analog Bits uses $\hat{x}_0 = f_{\theta}(x_0, \emptyset, t)$ as the approximation to $\tilde{x}_0$, which is applied on half of the samples for self-conditioning, while setting  the remaining samples to zero. Self-conditioning has been shown to remarkably improve the generation quality in practice and has been utilized in text diffusion models~\cite{DBLP:journals/corr/abs-2211-15089,DBLP:journals/corr/abs-2212-09412,DBLP:journals/corr/abs-2212-09462}. 
Following a similar motivation, Diffusion-NAT~\cite{zhou2023diffusionnat} proposes the self-prompting strategy, which takes the intermediate generated result as the prefix of the original condition text. Such a strategy can be repeated multiple times, reducing the inference difficulty at the early stage.

\paragraph{Semi-NAR Generation.} SSD-LM~\cite{DBLP:journals/corr/abs-2210-17432} introduces a semi-NAR decoding strategy. It generates a token block of size $B$ iteratively by concatenating a random noise with previously generated blocks as input. The generated block is then concatenated to the previous context to create a new context. The above process is repeated until the maximum desired length is reached. Such a generation strategy compensates for the lack of dependencies in the NAR generation of text diffusion models. 

\paragraph{Additional Normalization.}
It has been observed that rare tokens tend to have larger norms than common tokens~\cite{DBLP:journals/corr/abs-2212-09412}, while the noise scale added on different tokens is the same in existing diffusion models. Thus, rare tokens require more diffusion steps to be fully corrupted. To address this issue, Difformer introduces a layer normalization module to constrain the scale of word embeddings to the same level. 
\paragraph{Timestep Sampling.}
During training, most existing works randomly sample the timesteps and  the corresponding noise to corrupt the original data. However, existing studies~\cite{DBLP:conf/icml/NicholD21,DBLP:journals/corr/abs-2210-08933} find that sampling $t$ uniformly will add  noise to the objective function. Hence, they propose the importance timestep sampling: 
\begin{equation}
     \mathcal{L}_{\rm vlb}=\mathbb{E}_{t\sim p_t}\big[\frac{\mathcal{L}_t}{p_t}\big], p_t \propto\sqrt{\mathbb{E}[\mathcal{L}_t^2]}, \sum_{t=0}^{T-1}p_t=1.
\end{equation}
Since this method assigns higher weights to timesteps that incur larger losses, it has shown to be effective to stabilize the training process.

~\section{Conclusion and Future Directions} \label{sec: future directions}
This paper presents an overview of recent progress in text diffusion models. We review both the discrete and continuous text diffusion models, and discuss key designs in the diffusion process. 
We also summarize the utilization of PLMs in text diffusion models and introduce other optimization techniques for text data. 
To advance this field, there are several promising  directions for improving text diffusion models.

\paragraph{Customized Noise Schedules.}
Existing noise schedules in text diffusion models are mainly borrowed from image generation tasks, which generally treat all tokens equally in the forward or denoising process. As a result, they tend to neglect the difference between tokens in importance and frequency, causing  some tricky problems, \eg the inaccurate generation of keywords and rare words.
For this issue, 
DiffusionBERT~\cite{DBLP:journals/corr/abs-2211-15029} has proposed the spindle schedule, which assigns higher probabilities to corrupt more informative tokens and shows substantial performance improvement. 
More research can further explore text-tailored and task-specific noise schedules.

\paragraph{Efficient and Effective Utilization of  PLMs.} 
Although existing works~\cite{DBLP:journals/corr/abs-2211-15029,DBLP:journals/corr/abs-2212-09462} have successfully utilized PLMs in text diffusion models, they still struggle to surpass the fine-tuned performance of the original PLMs in text generation tasks. The reason is that these PLMs are pre-trained mainly for generating texts in a sequence-to-sequence or autoregressive manner, without the consideration of the diffusion process. It also raises another problem that existing works usually require more training steps to effectively adapt PLMs into the diffusion process. Therefore, it is important to investigate how to efficiently adapt PLMs into text diffusion models and effectively inspire their pre-learned rich knowledge for performing the denoising process.

\paragraph{Unified Multimodal Diffusion Models.}
Diffusion models have achieved remarkable  performance on text-to-image tasks~\cite{DBLP:journals/corr/abs-2204-06125}, and some recent studies  also pay attention to image-to-text generation such as image captioning~\cite{DBLP:journals/corr/abs-2212-03099}. 
Actually, the two lines of works generally adopt similar diffusion mechanisms or settings, but with different data formats (\ie text and image).
Therefore, by unifying the way of modeling text and image,  diffusion models are promising to unify and share the latent semantic spaces of both image-to-text and text-to-image generation tasks, achieving the goal of a one-for-all multimodal generation model.
There are several studies~\cite{DBLP:journals/corr/abs-2211-14842,DBLP:journals/corr/abs-2211-08332} that have explored this idea, and also discuss the possible issues about the unified diffusion model. 

\paragraph{Alignment with Human Values.}
As diffusion models own a strong capacity to generate diverse results, they may generate improper content that violates human values  \eg race or gender biases~\cite{DBLP:journals/corr/abs-2202-04053}.
This problem would be more serious once PLMs are also utilized in the diffusion model, since existing PLMs are pre-trained on large-scale corpora collected from the Internet, containing sensitive personal information or offensive sentences~\cite{DBLP:conf/emnlp/GehmanGSCS20}.
Considering the powerful abilities of text diffusion models to impose control on the intermediate results, more efforts should be devoted to developing their strong potential to prevent the above issues or steer the detoxification of the generated content~\cite{DBLP:conf/acl/LiuSLSBSC20}.

\section*{Acknowledgments}
This work was partially supported by National Natural Science Foundation of China under Grant No.
62222215, Beijing Natural Science Foundation under Grant No. 4222027, and Beijing Outstanding
Young Scientist Program under Grant No. BJJWZYJH012019100020098. And this work is also
partially supported by the Outstanding Innovative Talents Cultivation Funded Programs 2022 of
Renmin University of China. Xin Zhao is the corresponding author.

\bibliographystyle{named}
\bibliography{ijcai23}

\end{document}